%% file: main.tex
\documentclass[10pt, a4paper, logo]{googledeepmind}

\usepackage{times}
\usepackage{booktabs}
\usepackage{multirow}
\usepackage{graphicx}
\usepackage{subcaption}
\usepackage{wrapfig}
\usepackage{algorithm}
\usepackage{algorithmic}
\usepackage{tcolorbox}
\usepackage{enumitem}
\usepackage{microtype}
\usepackage{inconsolata}
\usepackage{amsmath}
\usepackage{amsfonts}
\usepackage{amssymb}
\usepackage{xcolor}
\usepackage{natbib}

\hypersetup{
  pdftitle={Generative Query Suggestion via Intent Coverage and Query-Level Credit Assignment},
  pdfauthor={Xinpeng Liu, Lu Ma, Jiayi Qiao, Mengyu Zhou, Linglong Li, Xiaofeng Bian, Haonan Chen, Xiaoxi Jiang, Guanjun Jiang}
}

\title{Generative Query Suggestion via Intent Coverage and Query-Level Credit Assignment}

\author[2]{Xinpeng Liu\textsuperscript{*,$\ddag$}}
\author[1]{Lu Ma\textsuperscript{$\ddag$}}
\author[3]{Jiayi Qiao\textsuperscript{*}}
\author[1]{Mengyu Zhou\textsuperscript{\dag}}
\author[4]{Linglong Li}
\author[1]{Xiaofeng Bian}
\author[1]{Haonan Chen}
\author[1]{Xiaoxi Jiang}
\author[1]{Guanjun Jiang}
\affil[1]{Qwen Business Unit of Alibaba}
\affil[2]{Peking University}
\affil[3]{National University of Singapore}
\affil[4]{Pengcheng National Laboratory}
\correspondingauthor{Mengyu Zhou (\texttt{zhoumengyu.zmy@alibaba-inc.com})}
\footnotetext{\textsuperscript{*}Work done during an internship at Alibaba.\quad
\textsuperscript{\dag}Corresponding author.\quad
\textsuperscript{$\ddag$}Equal contribution.}

\begin{abstract}
Generative query suggestion aims to enhance user engagement by anticipating user intents and recommending relevant follow-up queries. A central challenge is to generate slates whose individual queries are useful while the slate covers distinct intents. We propose an Intent-Driven Query Suggestion Framework with dual-stage optimization. First, \textit{intent-aware diversity modeling} constructs intent-aligned supervised fine-tuning (SFT) data and uses an Intent-Aware Diversity Reward to optimize intent coverage. Second, \textit{query-level credit assignment} routes individual quality signals to the corresponding query tokens while sharing a slate-level diversity signal across the slate. Experiments on a large-scale production dataset, including online A/B testing and offline evaluation, show improvements in click-through rate, query quality, and intent coverage.
\end{abstract}

\begin{document}
\maketitle

\section{Introduction}
\begin{wrapfigure}{r}{0.48\textwidth}
\vspace{-0.8\baselineskip}
\centering
\includegraphics[width=\linewidth]{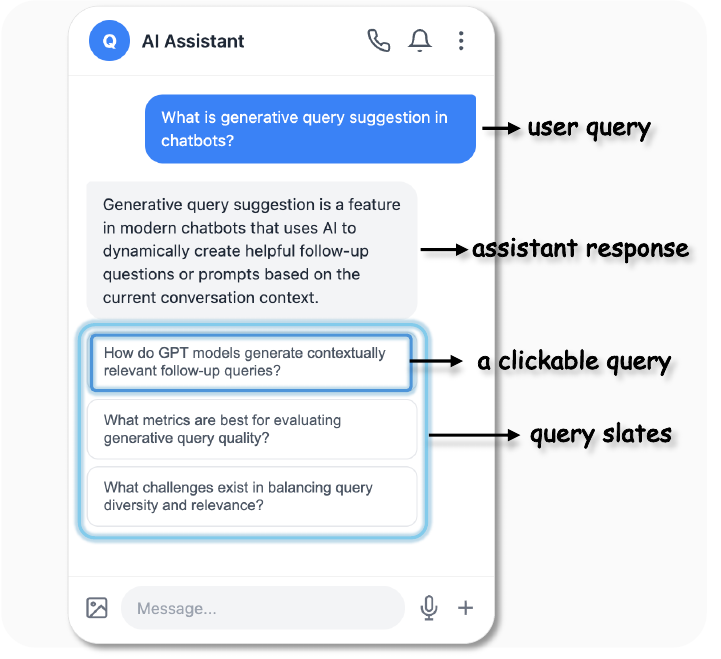}
\caption{Example of conversational query suggestion, where the AI Assistant recommends diverse follow-up queries to enhance user engagement.}
\label{fig:examples}
\vspace{-0.6\baselineskip}
\end{wrapfigure}
The rise of conversational AI assistants~\cite{achiam2023gpt} has fundamentally transformed how users access information and accomplish tasks. Recent work studies multimodal, click-guided, and preference-aligned query suggestion for conversational systems~\cite{wang2024multimodal,min2025ctr,yin2025clicks}. Figure~\ref{fig:examples} shows an example of conversational query suggestion, where diverse follow-up queries are recommended to enhance user engagement.
Recent advances~\cite{lin2023generative,bao2023tallrec} in generative query suggestion employ large language models (LLMs) to synthesize queries directly from conversational context. While generative query suggestion offers greater flexibility than retrieval-based methods, it requires dual optimization: query slates must be diverse to cover the breadth of potential user follow-up interests, while each individual query must remain contextually relevant to the conversation history, accurately capturing the user's underlying intent.

However, jointly optimizing these objectives presents two challenges. \textbf{L1) Intent-aware diversity optimization.} Semantic diversification through embedding similarity and post-hoc reranking is well established, but such methods do not necessarily optimize context-dependent coverage of distinct user intents during policy training. For instance, ``How to learn Python'' and ``Python tutorials'' are lexically distinct but express the same learning intent. We therefore model diversity as coverage over orthogonal intent dimensions rather than textual variation alone. \textbf{L2) Within-slate credit assignment.} \citet{min2025ctr} and \citet{yin2025clicks} align query suggestions with click- or preference-based objectives. Our method instead explicitly applies query-specific quality advantages only to the corresponding query tokens while sharing a slate-level intent-coverage advantage across the slate. This decomposition identifies each query's contribution within an autoregressively generated slate.

To address these challenges, we propose an \textbf{Intent-Driven Query Suggestion Framework} with dual-stage optimization for conversational AI assistants. To address L1, we introduce \textbf{intent-aware diversity modeling} in semantic intent space. During SFT, we construct a cold-start data pipeline based on user-initiative alignment. During reinforcement learning (RL), an Intent-Aware Diversity Reward (IAD-R) captures intent coverage through Noisy-OR aggregation. To address L2, we devise a \textbf{query-level credit assignment mechanism}: query-quality advantages are applied only to the corresponding query tokens, whereas a slate-level diversity advantage is shared across all queries. Their composite enables fine-grained optimization of individual quality while maintaining overall slate diversity.

We validate the framework through industrial online A/B testing and human evaluation. Its distinguishing feature is the combination of query-specific token-level advantages and a shared slate-level intent-coverage advantage in a single policy update.

\section{Related Work}

\paragraph{Generative Query Suggestion}
Large Language Models (LLMs) have been widely used to generate diverse and context-aware query suggestions~\cite{di2023retrieval}. Diversification can be imposed through decoding constraints~\cite{deng2025onerec,guo2025onesug}, embedding-based reranking~\cite{bacciu2024generating}, or preference alignment. \citet{min2025ctr} combine multi-source CTR modeling, diversity-aware CTR-weighted DPO, and iterative CTR calibration; \citet{yin2025clicks} introduce uncertainty-aware preference modeling, out-of-distribution regularization, and multi-stage alignment. These systems align suggestions with clicks or preferences. Our focus instead is the granularity of credit within a generated slate: query-specific quality advantages are routed only to the corresponding tokens, while intent coverage supplies a shared slate-level advantage. Inspired by exposure modeling in recommender systems~\cite{yin2025clicks,min2025ctr,chen2025unisearch}, we view each generated query as activating a user intent and optimize intent-space coverage directly.

\paragraph{Credit Assignment and Slate Optimization in RL}
Reinforcement Learning (RL) is commonly used to align generative models with human preferences~\cite{rafailov2023direct,ouyang2022training}, but slate-level rewards often lead to ambiguous credit assignment~\cite{yin2025clicks}. Prior studies address this with process reward modeling at the token or step level~\cite{yue2025promoting,xiong2025stepwiser,cui2025process}. SlateQ~\cite{ie2019slateq} decomposes slate Q-values into item-level scores over a fixed candidate pool, while our setting constructs slates autoregressively and optimizes them with policy gradients. We therefore introduce query-level credit assignment, which distributes slate rewards into per-query advantages and combines them with intent-aware diversity rewards.

\begin{figure*}[t]
\centering
\includegraphics[width=0.99\textwidth]{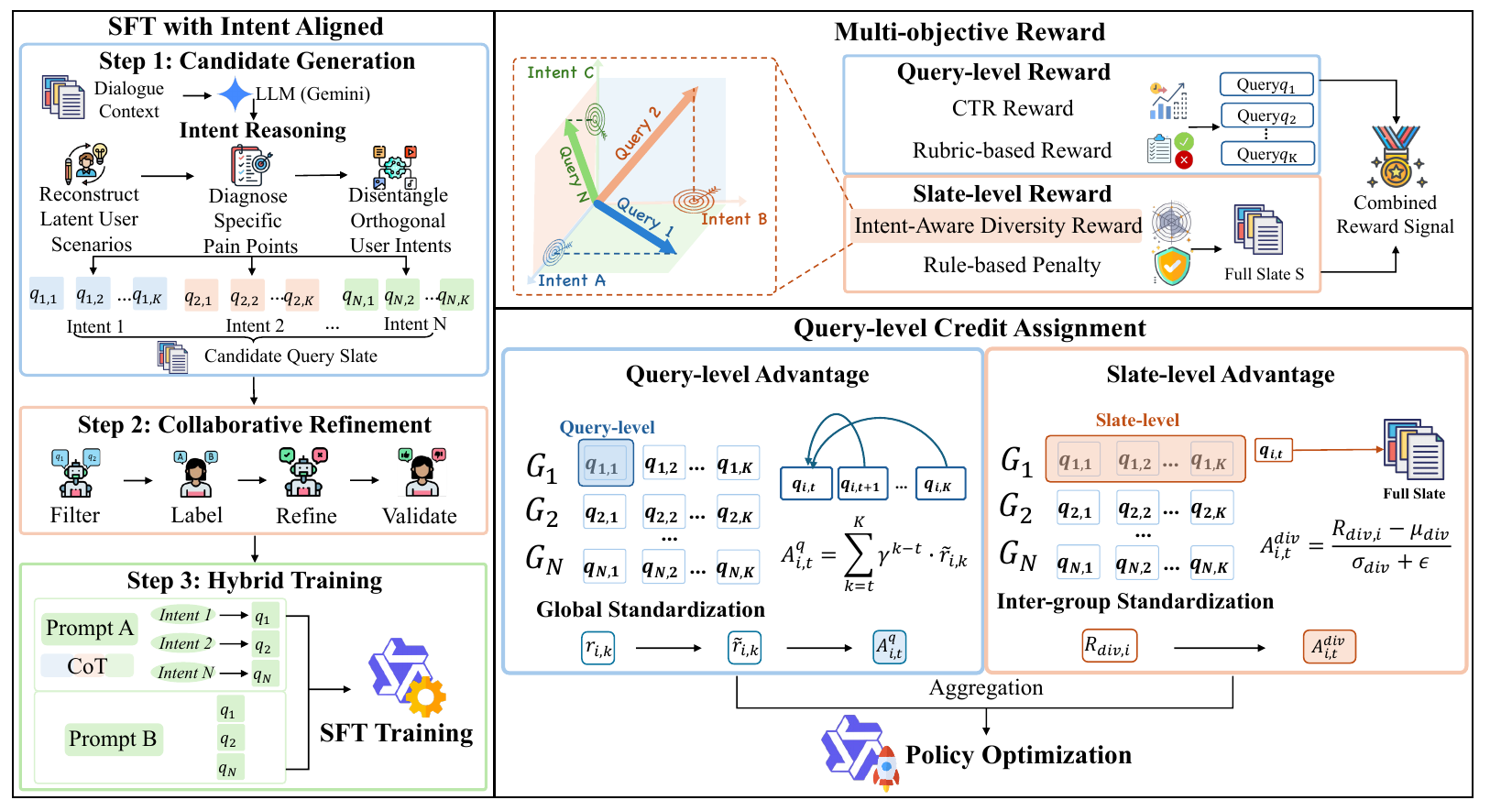}
\caption{Overview of our framework. Left: Cold-Start SFT stage with LLM generation, human-AI collaboration and hybrid CoT training. Right: RL stage with multi-objective reward and query-level credit assignment.}
\vspace{-10pt}
\label{fig:framework}
\end{figure*}

\section{Methodology}

We formalize conversational query suggestion as a slate generation task (\S\ref{sec:problem}) that jointly optimizes query-level quality and slate-level diversity. As illustrated in Figure~\ref{fig:framework}, our framework consists of two stages: (1) a cold-start supervised fine-tuning stage (\S\ref{sec:sft}) that constructs high-quality training data through LLM generation and human-AI collaboration, and (2) a reinforcement learning stage (\S\ref{sec:rl}) that optimizes intent-aware diversity and query-level credit assignment.

\subsection{Problem Formulation}
\label{sec:problem}

We formulate conversational query suggestion as a conditional generation task. Given context $\mathcal{X} = \{q_u, r, h\}$, where $q_u$ denotes the current user query, $r$ the assistant response, and $h$ the conversation history, the model generates a slate of $K$ candidate queries:
\begin{equation}
\mathcal{S} = \{q_1, \ldots, q_K\} \sim \pi_\theta(\cdot \mid \mathcal{X})
\end{equation}
where $\pi_\theta$ represents the policy model parameterized by $\theta$. The objective is to jointly optimize individual query quality and slate-level diversity.

\subsection{Cold-Start SFT}
\label{sec:sft}
To address the data quality and diversity bottleneck in the SFT, we design a two-stage data construction pipeline coupled with a hybrid training strategy.

\paragraph{Step 1: Candidate Generation.} 
We leverage LLMs to perform multi-dimensional intent analysis on dialogue context, decomposing user needs into orthogonal intent dimensions through scenario reconstruction and pain point diagnosis. For each identified intent, we generate a diverse set of candidate queries that address distinct user information needs.

\paragraph{Step 2: Collaborative Refinement.}
We adopt a multi-stage human-AI collaboration pipeline to ensure data quality. The LLM first filters the optimal query for each intent dimension. Human annotators and an LLM-based evaluator then iteratively label and cross-validate the samples, with disagreements surfaced for re-adjudication to reduce subjective variability. This process ensures both scalability and high annotation quality. Detailed annotation protocols, qualification criteria, and quality control procedures are provided in Appendix~\ref{app:annotation}.

\paragraph{Step 3: Hybrid Training.} 
To balance inference efficiency with reasoning capability, we construct a hybrid instruction dataset containing both Direct Query Generation and Intent-Aware Chain-of-Thought (CoT) samples. Each generation mode is guided by task-specific system prompts (representative excerpts appear in Appendix Figures~\ref{fig:direct_query_prompt} and~\ref{fig:intent_cot_prompt}). This hybrid approach enables the model to capture complex intent reasoning patterns while maintaining real-time generation throughput.

\subsection{Reinforcement Learning}
\label{sec:rl}

\subsubsection{Multi-Objective Reward}

\paragraph{CTR Reward}
To align suggestions with user preferences, we employ a lightweight BERT-based CTR model trained on historical click logs. Given conversational context $\mathcal{X}^{(\tau)} = \{q_u^{(\tau)}, r^{(\tau)}, h^{(\tau)}\}$ and candidate query $q_i$, the model predicts click probability:
\begin{equation}
R_{\text{CTR}}(q_i, \mathcal{X}^{(\tau)}) = \hat{p}(q_i \mid \mathcal{X}^{(\tau)}) \in [0, 1]
\end{equation}
This reward guides the policy toward generating user-preferred suggestions.

\paragraph{Rubric-based Reward}

To prevent reward models from exploiting surface-level patterns, we introduce \textit{Rubric-based Rewards} following~\cite{ye2023flask,liu2023gevalnlgevaluationusing}. We employ an LLM to assess whether each query satisfies expert-curated quality rubrics (summarized in Appendix Figure~\ref{fig:rubric_eval_prompt}). Given query $q$ and rubrics $\mathcal{R} = \{r_1, \dots, r_n\}$, the reward uses an all-pass criterion:
\begin{equation}
R_{\text{rubrics}}(q) = 
\begin{cases} 
1, & \text{if } \forall i, \, r_i(q) = 1 \\
0, & \text{otherwise}
\end{cases}
\end{equation}
where $r_i(q) \in \{0, 1\}$ indicates whether query $q$ satisfies the $i$-th rubric.
During RL training, Qwen-Max produces this rubric-based reward using the rubric structure summarized in Appendix Figure~\ref{fig:rubric_eval_prompt}.

\paragraph{Rule-based Penalty}
We impose rule-based constraints $R_{\text{rule}}(\hat{\mathcal{Q}}^{(t)}_{(\tau)})$ to ensure quality and prevent degenerate outputs: (1) redundancy penalty based on Jaccard similarity, (2) length penalty, and (3) format compliance penalty.

\paragraph{Intent-Aware Diversity Reward}
In deployed generative query suggestion, commonly used surface deduplication and post-hoc reranking do not explicitly optimize context-dependent intent coverage during training.

Given conversation history $h^{(\tau)}$, we leverage an LLM to infer $M$ orthogonal user intents $\mathcal{Z} = \{z_1, \dots, z_M\}$ during offline construction, verified by human annotators. We adopt a Noisy-OR model~\cite{pearl1988probabilistic} to measure slate $\mathcal{S}$'s coverage over the intent space. For intent $z_m$, the coverage probability represents the likelihood that at least one query in the slate activates this intent:
\begin{equation}
P_{\text{cov}}(z_m \mid \mathcal{S}) = 1 - \prod_{q_k \in \mathcal{S}} \left( 1 - \sigma\left(\text{sim}(q_k, z_m)\right) \right)
\end{equation}
where $\text{sim}(q_k, z_m) = \frac{\phi(q_k)^\top \phi(z_m)}{\|\phi(q_k)\|_2 \|\phi(z_m)\|_2}$ is the cosine similarity between query and intent embeddings from Sentence-BERT~\cite{reimers2019sentencebertsentenceembeddingsusing}, and $\sigma(\cdot)$ is the sigmoid function. The Intent-Aware Diversity Reward (IAD-R) averages coverage across all intents:
\begin{equation}
R_{\text{div}}(\mathcal{S}) = \frac{1}{M} \sum_{m=1}^{M} P_{\text{cov}}(z_m \mid \mathcal{S})
\end{equation}
We employ uniform weighting to ensure unbiased intent coverage. This prevents the model from prioritizing dominant intents at the expense of long-tail scenarios.

\subsubsection{Query-level Credit Assignment}
Our query-level credit assignment mechanism includes query-level quality advantages and slate-level diversity advantages.

\paragraph{Query-level Advantage Calculation}
For $G$ query slates $\{y_i\}_{i=1}^G$ sampled from policy $\pi_{\theta}$, where each slate consists of $K$ queries, we compute the query-level reward for the $k$-th query in the $i$-th slate as a weighted combination of rubric-based and CTR rewards:
\begin{equation}
r_{i,k} = \alpha \cdot \mathcal{R}_{\text{rubrics}}(q_{i,k}) + \beta \cdot \mathcal{R}_{\text{CTR}}(q_{i,k})
\end{equation}
Here, $\alpha=0.6$ weights rubric compliance and $\beta=0.4$ weights predicted CTR. The aggregated query-level reward ensures that the generated queries maintain semantic compliance and quality stability while maximizing actual click-through conversion efficiency.

In contrast to GRPO's group-level normalization over sequence returns, we perform query-level advantage estimation by globally normalizing rewards across all $G \times K$ queries:
\begin{equation}
\tilde{r}_{i,k} = \frac{r_{i,k} - \mu_r}{\sigma_r + \epsilon}
\end{equation}
where $\mu_r$ and $\sigma_r$ denote the mean and standard deviation computed over all sampled query rewards. To account for the sequential dependency between queries, we introduce a discount factor $\gamma \in [0, 1]$ and compute the advantage for the $t$-th query as:
\begin{equation}
A_{i,t}^{\text{q}} = \sum_{k=t}^{K} \gamma^{k-t} \cdot \tilde{r}_{i,k}
\end{equation}
This formulation values each query by both its immediate reward and its impact on subsequent generations, encouraging coherent and high-quality query sequences.

\paragraph{Slate-level Diversity Advantage}
To encourage diverse query lists, we introduce slate-level diversity advantages. As diversity is a collective property, we normalize the diversity reward $R_{\text{div}, i}$ within the group:
\begin{equation}
A_{i,t}^{\text{div}} = \frac{R_{\text{div}, i} - \mu_{\text{div}}}{\sigma_{\text{div}} + \epsilon}, \quad \forall t \in \{1, \dots, K\}
\end{equation}
where $\mu_{\text{div}}, \sigma_{\text{div}}$ are computed over $\{R_{\text{div}, j}\}_{j \in G}$. This advantage is uniformly assigned to all queries, encouraging complementary suggestions.

\begin{table*}[t]
\centering
\small
\begin{tabular}{lccccc}
\toprule
\textbf{Method} & \textbf{CTR Gain (\%)} $\uparrow$ & \textbf{GSB vs. base} $\uparrow$ & \textbf{LLM Critic} $\uparrow$ & \textbf{Self-BLEU} $\downarrow$ & \textbf{Intent Cov.} $\uparrow$ \\
\midrule
PE-based & +0 & +0 & 0.33 & 0.0656 & 0.27 \\
SFT-cold start~ & +27.06 & +0.16 & 0.67 & 0.0609 & 0.54 \\
PPO & +35.29 & +0.31 & 0.71 & 0.0571 & 0.69 \\
GRPO & +43.53 & +0.37 & 0.83 & 0.0474 & 0.85 \\
\midrule
\textbf{RL-Ours} & \textbf{+48.24} & \textbf{+0.44} & \textbf{0.88} & \textbf{0.0470} & \textbf{0.91} \\
\bottomrule
\end{tabular}
\caption{Main results. CTR reports relative gains over the PE-based arm; GSB is relative to PE-based outputs. Bold indicates the best result.}
\vspace{-10pt}
\label{tab:main_results}
\end{table*}

\paragraph{Advantage Aggregation and Policy Optimization}
We combine query quality and slate diversity through a composite advantage with rule-based gating:
\begin{equation}
A_{i,t} = 
\begin{cases} 
\frac{A_{i,t}^{\text{q}} + A_{i,t}^{\text{div}}}{\sqrt{2}}, & \text{if } \text{Rule}(y_i) = \text{Pass} \\
-1, & \text{if } \text{Rule}(y_i) = \text{Fail}
\end{cases}
\end{equation}
where quality and diversity advantages are equally weighted and normalized. Slates violating critical constraints receive $-1$ penalty, combining learned and rule-based rewards.

Following GRPO, we optimize token-level generation probabilities using query-level advantages. To bridge this granularity gap, we uniformly broadcast each query-level advantage to all its tokens:
\begin{equation}
A^{\text{tok}}_{j} = A_{i,t}, \quad \forall u_{j} \in \text{tokens}(q_{i,t})
\end{equation}

The policy is then optimized via the clipped surrogate objective:
{\small
\begin{equation}
\mathcal{L}(\theta)
= -\mathbb{E}_{y \sim \pi_\theta(\cdot \mid x)}
\left[
\begin{aligned}
&\sum_j \min\!\left(\rho_j A_j^{\text{tok}},\,
\bar{\rho}_j A_j^{\text{tok}}\right) \\
&\quad - \lambda_{\text{KL}}D_{\text{KL}}
\!\left(\pi_\theta \| \pi_{\text{ref}}\right)
\end{aligned}
\right].
\end{equation}
}
where $\rho_j = \pi_{\theta}(u_j | u_{<j}) / \pi_{\text{ref}}(u_j | u_{<j})$ is the probability ratio, $\bar{\rho}_j = \text{clip}(\rho_j, 1-\epsilon, 1+\epsilon)$, and $\lambda_{\text{KL}}$ is the KL coefficient. The clipping mechanism prevents large policy updates, while the KL penalty ensures training stability.

\section{Experiments}

\subsection{Experimental Setup}
We evaluate on a large-scale production system using anonymized real-world interactions. We use Qwen3-30B-A3B~\cite{qwen3} as the foundation model (training details in Appendix~\ref{sec:training_configuration}).

\vspace{-5pt}
\paragraph{Training Datasets.} We collect data from real user logs of a large-scale conversational AI assistant. For SFT, we construct 15,000 instances from 4 weeks of conversations using the pipeline in \S\ref{sec:sft}, mixing 20\% intent-augmented CoT samples with 80\% direct generation samples. For RL, we sample 12,000 instances from an additional 2 weeks, annotated with intents for IAD-R.
\vspace{-5pt}
\paragraph{Test Datasets.} We construct two expert-validated test sets spanning diverse user scenarios. The \textit{Offline Evaluation Set} contains 500 contexts for automatic metrics. The \textit{Human Evaluation Set} contains 150 representative contexts for measuring generation quality and diversity.

\vspace{-5pt}
\paragraph{Online Metrics.} To assess real-world efficacy, we conducted a one-week A/B test on live traffic. The core metric is \textit{Click-Through Rate} (CTR), which measures interaction with the suggestion slate within a single turn. The five systems ran concurrently in mutually exclusive, randomly assigned user-level buckets. Traffic allocation for PE-based/SFT/PPO/GRPO/RL-Ours was 5/80/5/5/5\%; SFT was the incumbent production arm, while Table~\ref{tab:main_results} reports CTR gains relative to the PE-based arm. We use two-proportion $z$-tests on unrounded arm-level CTRs. Exact traffic counts and absolute CTR values are commercially sensitive.
\vspace{-5pt}
\paragraph{Offline Metrics.} We evaluate query quality with: (1) \textit{Good-Same-Bad} (GSB), a pairwise side-by-side human evaluation across content quality, text quality, intent relevance, and information gain (Table~\ref{tab:human_eval_standards}), yielding $(\#\text{Good} - \#\text{Bad})/N$; and (2) \textit{LLM Critic}, an ensemble of Gemini-3, Claude-4.5, and Qwen-Max that reports the majority-voted checklist pass rate (Figure~\ref{fig:rubric_eval_prompt}). Thus, no single Qwen-family evaluator determines this metric. For slate diversity, we use: (1) \textit{Self-BLEU}~\cite{yin2025clicks,min2025ctr}, the average pairwise similarity within each slate; and (2) \textit{Intent Coverage}, the recall of separately annotated test intents, where three annotators independently label intent matches for each query and majority voting determines the final label.

Intent Coverage targets the same property as IAD-R, and LLM Critic uses the same rubric family as the training reward. We therefore treat both as target-aligned diagnostics rather than fully independent evidence. Neither evaluation directly reuses stored training-time reward outputs: evaluation coverage uses discrete human matching against separately annotated test intents instead of Sentence-BERT similarity and Noisy-OR aggregation, while LLM Critic recomputes judgments on held-out contexts using the three-model ensemble. Because Qwen-Max participates in both training and the evaluation ensemble, online CTR and human GSB serve as the primary independent evidence.

\subsection{Overall Results and Analysis}
\label{sec:main_results}

Table~\ref{tab:main_results} compares our RL framework with the production and policy-optimization baselines. Scalability to Qwen3-235B-A22B is evaluated in Appendix~\ref{app:scalability}.

\textbf{SFT Establishes a Strong Foundation.} Starting from the PE-based prompt engineering baseline, supervised fine-tuning on user interaction data improves all five metrics. As shown in Table~\ref{tab:main_results}, the SFT-cold start strategy achieves a +27.06\% CTR improvement, a GSB score of +0.16, and an LLM Critic score of 0.67. These results support the value of learning from real user interactions beyond prompt engineering.

\textbf{RL Methods Improve Performance.} We explore PPO and GRPO using the same reward signals as our method. PPO increases the PE-relative CTR gain to +35.29\%. GRPO reaches +43.53\% CTR gain, +0.37 GSB, and 0.83 LLM Critic. Relative to the incumbent SFT arm, GRPO improves CTR by 13.0\% ($p = 0.006$).

\textbf{Query-level Credit Assignment Achieves the Best Results.} Our method attains the highest PE-relative CTR gain (+48.24\%). The 4.71 difference from GRPO is measured in percentage points of PE-relative gain; the direct comparison is a 3.25\% relative CTR improvement ($p = 0.02$). RL-Ours also improves CTR over the incumbent SFT arm by 16.7\% ($p = 0.003$). On the common 150-context paired evaluation subset, two-sided paired $t$-tests comparing RL-Ours with GRPO give GSB +0.07 ($p = 0.03$), Intent Coverage +0.06 ($p = 0.002$), and LLM Critic +0.05 ($p = 0.05$). Self-BLEU changes from 0.0474 to 0.0470, whereas Intent Coverage increases from 0.85 to 0.91. Because Intent Coverage and LLM Critic are aligned with their corresponding training objectives, CTR and human GSB are the primary independent evidence for the improvement.

\begin{table}[t]
\centering
\begin{minipage}[t]{0.48\textwidth}
\centering
\scriptsize
\setlength{\tabcolsep}{2.5pt}
\resizebox{\linewidth}{!}{%
\begin{tabular}{lcccc}
\toprule
\textbf{Method} & \shortstack{\textbf{GSB vs.}\\\textbf{base}} & \shortstack{\textbf{LLM}\\\textbf{Critic}} & \textbf{Self-BLEU} & \shortstack{\textbf{Intent}\\\textbf{Cov.}} \\
\midrule
PE-based & +0 & 0.33 & 0.0656 & 0.27 \\
SFT-Candidate & +0.05 & 0.41 & 0.0631 & 0.28 \\
SFT-Refined & +0.13 & 0.59 & 0.0617 & 0.42 \\
SFT-Hybrid & \textbf{+0.16} & \textbf{0.67} & \textbf{0.0609} & \textbf{0.54} \\
\bottomrule
\end{tabular}%
}
\caption{Ablation study on data pipeline stages. Self-BLEU $\downarrow$, other metrics $\uparrow$.}
\label{tab:ablation_sft}
\end{minipage}
\hfill
\begin{minipage}[t]{0.48\textwidth}
\centering
\scriptsize
\setlength{\tabcolsep}{2.5pt}
\resizebox{\linewidth}{!}{%
\begin{tabular}{lcccc}
\toprule
\textbf{Method} & \shortstack{\textbf{$\Delta$GSB vs.}\\\textbf{RL-Ours}} & \shortstack{\textbf{LLM}\\\textbf{Critic}} & \textbf{Self-BLEU} & \shortstack{\textbf{Intent}\\\textbf{Cov.}} \\
\midrule
\textbf{RL-Ours} & \textbf{0.00} & \textbf{0.88} & \textbf{0.0470} & \textbf{0.91} \\
w/o IAD-R & $-0.12$ & 0.83 & 0.0461 & 0.71 \\
w/o Rubric & $-0.09$ & 0.62 & 0.0496 & 0.86 \\
w/o CTR & $-0.11$ & 0.76 & 0.0486 & 0.89 \\
w/o query-level & $-0.07$ & 0.83 & 0.0474 & 0.85 \\
\bottomrule
\end{tabular}%
}
\caption{Ablation on RL components. GSB is measured relative to RL-Ours; the other columns report absolute values. ``query-level'' refers to query-level credit assignment. Self-BLEU $\downarrow$, other metrics $\uparrow$.}
\label{tab:ablation_rl}
\end{minipage}
\vspace{-10pt}
\end{table}

\begin{figure}[t]
\centering
\begin{subfigure}{0.49\linewidth}
    \centering
    \includegraphics[width=\linewidth]{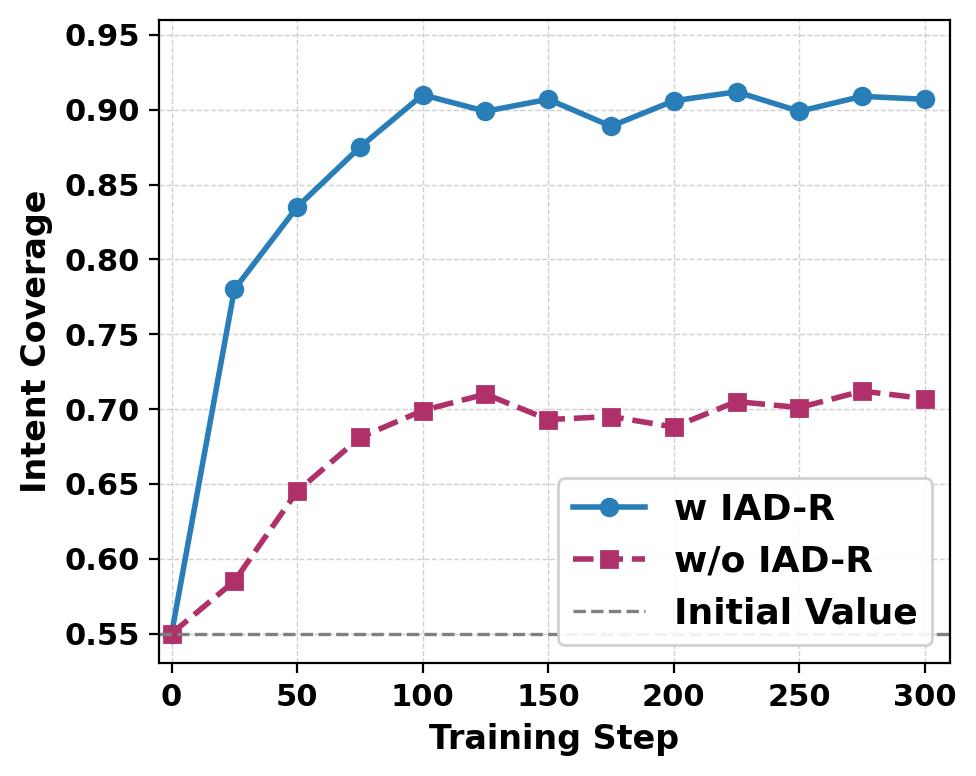}
    \caption{IAD-R}
    \label{fig:training_curves_a}
\end{subfigure}
\hfill
\begin{subfigure}{0.49\linewidth}
    \centering
    \includegraphics[width=\linewidth]{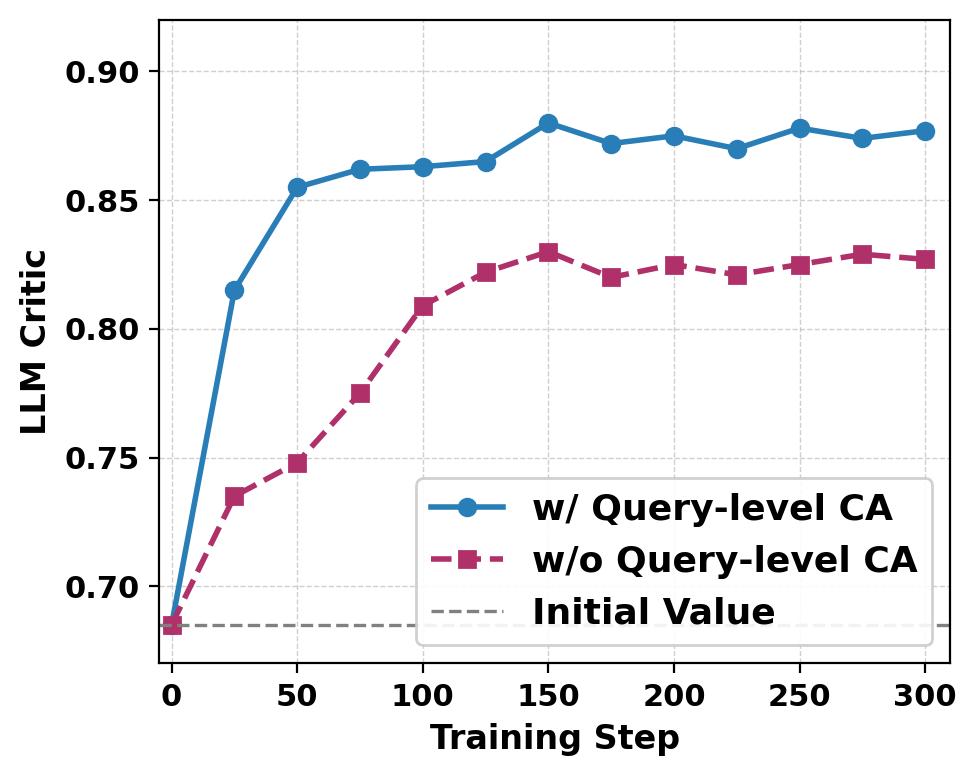}
    \caption{Query-level CA}
    \label{fig:training_curves_b}
\end{subfigure}
\vspace{-4pt}
\caption{Training dynamics of ablated components. (a) IAD-R impact on Intent Coverage. (b) Query-level credit assignment on LLM Critic.}
\vspace{-15pt}
\label{fig:training_curves}
\end{figure}

\subsection{Ablation Studies}
\label{sec:ablation_sft}

\paragraph{Ablation Study on SFT Data.}

To validate our data construction pipeline (Section~\ref{sec:sft}), we conduct an ablation study using different data sources: \textit{PE-based} uses only prompt engineering without fine-tuning, \textit{SFT-Candidate} trains on LLM-generated candidates (Stage 1), \textit{SFT-Refined} trains on human-AI collaborative refined queries (Stage 2), and \textit{SFT-Hybrid} trains on the full pipeline with mixed CoT traces and queries (Stage 3).

Table~\ref{tab:ablation_sft} shows progressive improvements across pipeline stages. Relative to SFT-Candidate, SFT-Refined improves GSB by 0.08 and Intent Coverage by 0.14, supporting the value of human-AI refinement. SFT-Hybrid adds a further 0.12 in Intent Coverage, suggesting that mixing CoT traces with direct queries helps the model learn intent-aware generation. We adopt SFT-Hybrid as the cold-start model for RL training.

\paragraph{Ablation Study on RL Components.}

We systematically remove key components from our RL framework to validate 
their individual contributions. Table~\ref{tab:ablation_rl} presents the 
final performance, while Figure~\ref{fig:training_curves} illustrates the 
training dynamics to provide deeper insights into how these components affect 
the learning process.

Removing diversity reward causes the largest Intent Coverage drop, decreasing 
from 0.91 to 0.71. Figure~\ref{fig:training_curves}(a) 
shows models with this reward converge faster and to a higher plateau, reaching 
0.91 by step 100, whereas the ablated variant continues to improve gradually, 
only approaching 0.71 by step 125. Both quality reward components prove 
essential. Without rubric-based reward, the LLM Critic score drops 
from 0.88 to 0.62, showing its role in maintaining generation 
quality. Similarly, removing CTR reward decreases GSB by 0.11, suggesting 
that engagement signals effectively guide the model toward user-preferred 
queries.

Query-level credit assignment also outperforms 
the sequence-level baseline. Figure~\ref{fig:training_curves}(b) demonstrates 
consistent advantages throughout training. As early as step 50, our method 
achieves an LLM Critic score of approximately 0.86 compared to 0.75 for the 
baseline, maintaining a consistent gap through convergence with final scores 
of 0.88 and 0.83 respectively. This result supports the conclusion that assigning rewards at 
the query level rather than the sequence level enables more effective optimization 
by providing precise feedback for each individual query.

\section{Conclusion}
We propose an intent-driven framework that jointly optimizes query quality and slate diversity through dual-stage optimization. The online and offline results show that query-level credit assignment improves engagement and human-evaluated quality while intent-aware training increases semantic coverage. Future work will explore personalization and multimodal query generation.

\section*{Limitations}
This work has several limitations. First, our training and online evaluation are based on anonymized production logs from a deployed conversational assistant; although this setting provides realistic feedback, the proprietary nature of the data limits direct reproducibility. We therefore report detailed data construction and evaluation protocols and plan to release a sanitized evaluation subset when compliance permits. Second, online CTR is an engagement proxy rather than a direct measure of user satisfaction or task success. We interpret it together with offline human evaluation, and future work should incorporate richer satisfaction and task-completion signals. The online analysis reports significance from a single one-week experiment but does not report confidence intervals or repeated-window variance. Third, Intent Coverage targets the same property as IAD-R, and LLM Critic uses the same rubric family as the training reward; Qwen-Max also participates in both training and the evaluation ensemble. The metrics do not reuse stored training-time reward outputs, but they are not fully independent evidence, so our main conclusions foreground online CTR and human GSB. Fourth, our evaluation uses independent annotations and majority voting but does not report a chance-corrected inter-annotator agreement statistic. Our intent-aware diversity reward also relies on LLM-assisted intent construction and human validation; remaining annotation noise or intent granularity mismatch may affect coverage estimation. Repeated full-scale RL runs and a complete sweep over the quality--diversity weights are computationally prohibitive, so we report evaluation-sample significance tests and only a targeted discount-factor sensitivity study. We also do not measure whether broader intent coverage improves multi-turn engagement conditioned on distinct follow-up intents. Finally, while query-level advantages improve optimization granularity, diversity remains a slate-level property shared across its queries; more precise marginal contribution estimation is an important direction for future work.

\section*{Ethics Statement}
\paragraph{Use of AI Assistants}
We used ChatGPT solely as a writing assistant to polish the manuscript after the initial draft was written. The authors were responsible for all scientific content, analyses, and conclusions presented in this paper.

\bibliography{references}
\bibliographystyle{conference}

\clearpage 
\appendix
\raggedbottom

\section{Prompt Templates}
\label{sec:prompts}
We provide abridged prompt excerpts that illustrate the formats used for direct query generation (Figure~\ref{fig:direct_query_prompt}), intent-guided CoT reasoning (Figure~\ref{fig:intent_cot_prompt}), and rubric-based evaluation (Figure~\ref{fig:rubric_eval_prompt}). Some operational rules are omitted for brevity.
\input{prompts/DirectQuery}
\input{prompts/IntentCoT}
\input{prompts/LLMjudge}

\section{Scalability to Larger Models}
\label{app:scalability}

\begin{wraptable}{r}{0.48\textwidth}
\vspace{-0.8\baselineskip}
\centering
\scriptsize
\setlength{\tabcolsep}{2.5pt}
\resizebox{\linewidth}{!}{%
\begin{tabular}{lcccc}
\toprule
\textbf{Method} & \shortstack{\textbf{GSB vs.}\\\textbf{base}} & \shortstack{\textbf{LLM}\\\textbf{Critic}} & \textbf{Self-BLEU} & \shortstack{\textbf{Intent}\\\textbf{Cov.}} \\
\midrule
PE-based & +0 & 0.57 & 0.0493 & 0.58 \\
SFT-cold start  & +0.14 & 0.79 & 0.0471 & 0.76 \\
\textbf{RL-Ours} & \textbf{+0.23} & \textbf{0.92} & \textbf{0.0463} & \textbf{0.94} \\
\bottomrule
\end{tabular}%
}
\caption{Scalability results on Qwen3-235B-A22B. Self-BLEU $\downarrow$, other metrics $\uparrow$.}
\label{tab:scalability_235b}
\vspace{-0.5\baselineskip}
\end{wraptable}

To validate the generalizability of our framework across model scales, we conduct experiments on Qwen3-235B-A22B using offline metrics due to deployment constraints. As shown in Table~\ref{tab:scalability_235b}, our approach achieves 0.92 LLM Critic and 0.94 Intent Coverage compared with the baseline's 0.57 and 0.58. Despite the larger model's stronger baseline (0.57 vs. 0.33 LLM Critic on Qwen3-30B-A3B), our framework retains improvements of 61\% in LLM Critic and 62\% in Intent Coverage. Self-BLEU decreases from 0.0493 to 0.0471 and 0.0463, showing that SFT data construction and RL optimization also reduce lexical redundancy at the larger model scale.

\begin{table}[ht]
\centering
\begin{minipage}[t]{0.43\textwidth}
\centering
\scriptsize
\resizebox{\linewidth}{!}{%
\begin{tabular}{lccc}
\toprule
\textbf{Parameter} & \textbf{Value} & \textbf{Parameter} & \textbf{Value} \\
\midrule
Optimizer & AdamW & LR Scheduler & Linear \\
Learning Rate & 1e-5 & Grad Clip & 1.0 \\
$\beta_1$ & 0.99 & $\beta_2$ & 0.999 \\
Batch Size & 128 & Max Tokens & 8192 \\
\bottomrule
\end{tabular}%
}
\captionof{table}{Hyper-parameters for the SFT stage.}
\label{tab:sft_params}
\end{minipage}
\hfill
\begin{minipage}[t]{0.55\textwidth}
\centering
\scriptsize
\resizebox{\linewidth}{!}{%
\begin{tabular}{lccc}
\toprule
\textbf{Parameter} & \textbf{Value} & \textbf{Parameter} & \textbf{Value} \\
\midrule
Optimizer & AdamW & LR Schedule & Linear \\
LR & 2e-6 & Grad Clip & 1.0 \\
$\beta_1$ & 0.99 & $\beta_2$ & 0.999 \\
Batch Size & 64 & Max Tokens & 9216 \\
$\lambda_{\text{KL}}$ & 0.05 & Rollout $N$ & 8 \\
WD & 0.1 & Warmup & 2 \\
Clip Ratio & [0.2, 0.28] & Temp. & 1.0 \\
Top-$p$ & 1.0 & Epochs & 2 \\
Rubric wt. $\alpha$ & 0.6 & CTR wt. $\beta$ & 0.4 \\
$\gamma$ & 0.95 & Slate Size $K$ & 3 \\
\bottomrule
\end{tabular}%
}
\captionof{table}{Hyper-parameters for the RL stage.}
\label{tab:rl_params}
\end{minipage}
\end{table}

We use $\gamma = 0.95$ and fix $K = 3$ because the production interface displays three suggestions; neither value was tuned on the test set. The rubric/CTR mixture $(\alpha,\beta)=(0.6,0.4)$ is fixed across experiments. Table~\ref{tab:gamma_sweep} reports a targeted offline sensitivity analysis with all other settings fixed. Both metrics remain within a narrow range. Intent Coverage requires per-model human annotation and is therefore reported only for the selected $\gamma = 0.95$ setting (0.91 in Table~\ref{tab:main_results}).

\begin{table}[ht]
\centering
\small
\begin{tabular}{lcccc}
\toprule
$\gamma$ & 0.90 & \textbf{0.95} & 0.99 & 1.00 \\
\midrule
LLM Critic $\uparrow$ & 0.871 & \textbf{0.880} & 0.868 & 0.873 \\
Self-BLEU $\downarrow$ & 0.0491 & \textbf{0.0470} & 0.0482 & 0.0479 \\
\bottomrule
\end{tabular}
\caption{Offline sensitivity to the query-level discount factor $\gamma$.}
\label{tab:gamma_sweep}
\end{table}

\section{Annotation Pipeline and Quality Control}
\label{app:annotation}

\paragraph{Annotation Pipeline.}
The collaborative refinement in Step~2 follows a four-stage procedure:
(i) the LLM filters the optimal query for each intent dimension;
(ii) human annotators verify and label a seed subset of samples to establish quality standards;
(iii) an LLM-based evaluator performs batch labeling guided by human feedback;
(iv) human validators review and refine the LLM-labeled results.

\paragraph{Annotator Training and Qualification.}
All annotators follow standardized guidelines and undergo rigorous training. Qualification requires achieving \textgreater90\% agreement in back-to-back validation against gold-standard samples before being assigned to production labeling.

\paragraph{Quality Control.}
An independent QC team audits annotation quality on sampled batches. In addition, we apply machine-assisted cross-validation between LLM and human labels: samples with disagreement are automatically flagged and routed for re-adjudication. This mechanism reduces variability in subjective decisions and provides a continuous feedback loop for guideline refinement.

\section{Training Configuration}
\label{sec:training_configuration}
Model training is conducted using a cluster of 256 NVIDIA H100 GPUs. The specific training parameters for each stage are summarized in Tables~\ref{tab:sft_params} and~\ref{tab:rl_params}.

\section{Algorithm Details}
\label{appendix:algorithm}

The complete pseudocode of our query-level credit assignment mechanism is provided in Algorithm~\ref{alg:query_wise_credit}.

\begin{algorithm}[htbp]
\small
\caption{Query-level Credit Assignment}
\label{alg:query_wise_credit}
\begin{algorithmic}[1]
\REQUIRE Policy $\pi_{\theta}$, ref. policy $\pi_{\text{ref}}$, input $x$
\REQUIRE Hyperparams: $G$, $K$, $\alpha$, $\beta$, $\gamma$, $\epsilon$, $\lambda_{\text{KL}}$
\ENSURE Updated policy parameters $\theta$

\STATE \textbf{Step 1: Sample Generation}
\FOR{$i = 1$ \TO $G$}
    \STATE Sample slate $y_i = \{q_{i,1}, \ldots, q_{i,K}\} \sim \pi_{\theta}(\cdot \mid x)$
\ENDFOR

\STATE \textbf{Step 2: Query-level Reward}
\STATE Compute $r_{i,k} = \alpha \cdot \mathcal{R}_{\text{rubrics}}(q_{i,k}) + \beta \cdot \mathcal{R}_{\text{CTR}}(q_{i,k})$

\STATE \textbf{Step 3: Query-level Advantage}
\STATE Normalize: $\tilde{r}_{i,k} = \frac{r_{i,k} - \mu_{r}}{\sigma_{r} + \epsilon}$
\STATE Compute: $A_{i,t}^{\text{q}} = \sum_{k=t}^{K} \gamma^{k-t} \cdot \tilde{r}_{i,k}$

\STATE \textbf{Step 4: Diversity Advantage}
\STATE Compute $R_{\text{div}, i} = \mathcal{R}_{\text{div}}(y_i)$
\STATE Normalize: $A_{i,t}^{\text{div}} = \frac{R_{\text{div}, i} - \mu_{\text{div}}}{\sigma_{\text{div}} + \epsilon}$

\STATE \textbf{Step 5: Advantage Aggregation}
\STATE $A_{i,t} = \begin{cases} (A_{i,t}^{\text{q}} + A_{i,t}^{\text{div}})/\sqrt{2}, & \text{Pass} \\ -1, & \text{Fail} \end{cases}$

\STATE \textbf{Step 6: Token-level Broadcasting}
\STATE Assign $A^{\text{tok}}_j = A_{i,t}$ for all $u_j \in \text{tokens}(q_{i,t})$

\STATE \textbf{Step 7: Policy Optimization}
\STATE Compute $\rho_j = \frac{\pi_{\theta}(u_j \mid u_{<j})}{\pi_{\text{ref}}(u_j \mid u_{<j})}$
\STATE $\mathcal{J} = \sum_{j} \min(\rho_j A^{\text{tok}}_j, \bar{\rho}_j A^{\text{tok}}_j) - \lambda_{\text{KL}}D_{\text{KL}}(\pi_\theta\|\pi_{\text{ref}})$
\STATE $\mathcal{L}(\theta) = -\mathbb{E}[\mathcal{J}]$
\STATE Update: $\theta \leftarrow \theta - \nabla_{\theta} \mathcal{L}(\theta)$

\RETURN Updated policy $\pi_{\theta}$
\end{algorithmic}
\end{algorithm}

\section{Case study}

\begin{figure*}[t]
\centering
\includegraphics[width=0.87\textwidth]{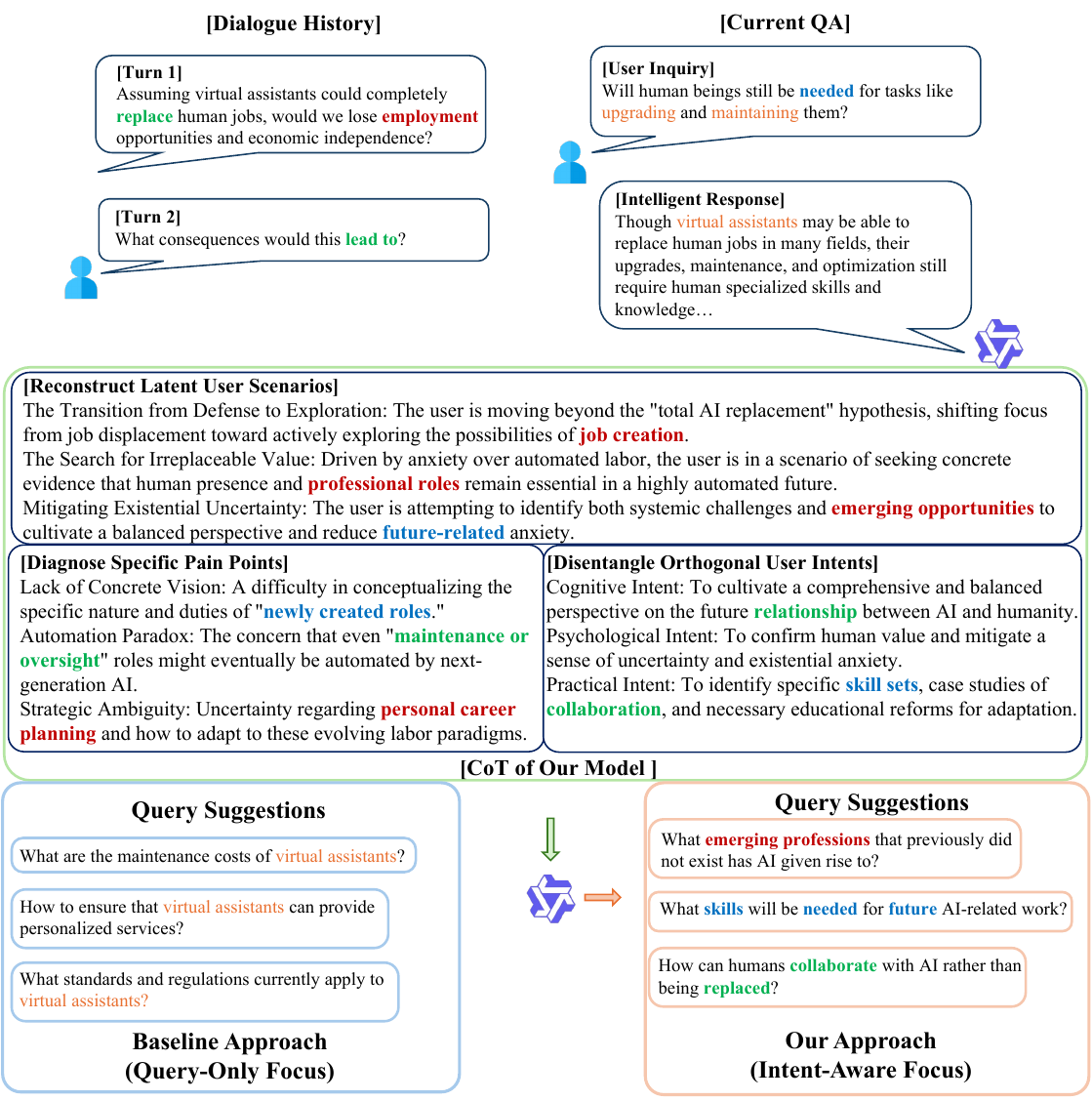}
\caption{Case study comparing intent modeling approaches between baseline and our model.}
\label{fig:case_study}
\end{figure*}

Traditional query suggestion models focus solely on the current turn, providing surface-level recommendations while neglecting global dialogue context and users' implicit needs. In contrast, our model integrates dialogue history with the current QA trajectory to perform multi-dimensional intent modeling across multiple facets: \textbf{Reconstructing Latent User Scenarios}, \textbf{Diagnosing Specific Pain Points}, and \textbf{Disentangling Orthogonal User Intents}.

As illustrated in Figure~\ref{fig:case_study}, the baseline captures only surface-level intent, while our model performs implicit need inference and intent-driven analysis. The generated suggestions maintain contextual continuity and precisely address users' latent intentions.

\section{Offline Evaluation Standard}
\label{appendix:eval_standard}

To systematically assess the contextual quality of the generated query suggestions, we establish a manual offline evaluation standard. The scoring protocol operates on a per-case basis, where each test case consists of three query suggestions generated from the preceding context. The score assigned to each case directly corresponds to the number of suggestions deemed valuable by the annotator. Table~\ref{tab:human_eval_standards} presents the detailed evaluation criteria across four dimensions: Content Quality, Requirement Satisfaction, Text Quality, and Information Gain.

\begin{table*}[t]
\centering
\small
\setlength{\tabcolsep}{3.5pt}
\begin{tabular}{p{3cm}p{12.5cm}}
\toprule
\textbf{Criterion} & \textbf{Description} \\
\midrule
\multicolumn{2}{l}{\textbf{Content Quality}} \\
\midrule
Timeliness Issue & The recommended query is outdated and no longer meets current requirements. \\
Security \& Compliance & Content involving pornography, politics, violence, or illegal activities is deemed non-compliant. \\
Question Duplication & Recommended queries should avoid significant (over 80\%) textual duplication. \\
History Duplicate & Overlapping less than 50\% with historical queries, offering additional information gain. \\
Duplicate-of-Response & Overlapping less than 50\% with previous responses and providing additional information. \\
\midrule
\multicolumn{2}{l}{\textbf{Requirement Satisfaction}} \\
\midrule
Irrelevant & The query severely deviates from the user's current primary needs and intent. \\
Over-Assumption & Makes unfounded assumptions about the user's identity, context, or expertise. \\
Over-Reliance & Overly focuses on past conversations, neglecting current needs. Prioritize the current round. \\
\midrule
\multicolumn{2}{l}{\textbf{Text Quality}} \\
\midrule
Word-Count-Exceeded & Exceeds 20 characters (Chinese, English x2, digits, punctuation, spaces each count as 1). \\
Unit Error & Non-standard units not commonly used, increasing user comprehension difficulty. \\
Poorly-Expression & The text is unclear, poorly expressed, with grammatical errors, typos, or ambiguities. \\
Inappropriate Topic & Pose questions about trending topics conversationally---avoid simply listing news. \\
\midrule
\multicolumn{2}{l}{\textbf{Information Gain}} \\
\midrule
Logical Confusion & The query is logically inconsistent or contradictory with the preceding content. \\
Generic Question & The query is too vague, generic, and common knowledge, offering little added value. \\
\bottomrule
\end{tabular}
\caption{Human evaluation standards for query suggestion quality assessment.}
\label{tab:human_eval_standards}
\end{table*}

\end{document}

%% file: prompts/DirectQuery.tex
\begin{figure}[ht]
\centering
\small
\begin{tcolorbox}[
  title= System Prompt for Direct Query Generation,
  colframe=gray!70!black,
  colback=gray!5
]
\textbf{\# Role} \\
User Intent Insight Expert tasked with generating follow-up queries on behalf of users.

\vspace{1mm}
\textbf{\# Workflow}

\textbf{1. Intent Categorization}
\begin{itemize}[itemsep=1pt, topsep=2pt]
    \item \textbf{Creative}: Content generation/refinement tasks (writing, coding, translation).
    \item \textbf{Non-Creative}: Factual Q\&A, information retrieval, troubleshooting.
\end{itemize}

\textbf{2. Strategy}
\begin{itemize}[itemsep=1pt, topsep=2pt]
    \item \textbf{Creative}: Generate instructive queries (style, audience, tone optimization).
    \item \textbf{Non-Creative}: Explore extended value and practical needs.
\end{itemize}

\textbf{\# Constraints}
\begin{itemize}[itemsep=1pt, topsep=2pt]
    \item Generate 3-5 diverse queries (no duplicates or highly similar ones).
    \item Character limit: 20 characters or fewer (preferably 8-16).
    \item Avoid repeating historical queries or answered content.
\end{itemize}

\hrulefill

\textbf{Example:} \\
What scenarios suit deep learning? \\
How does ML do feature engineering? \\
What GPUs for training models?
\end{tcolorbox}
\caption{Abridged system prompt for direct query generation.}
\label{fig:direct_query_prompt}
\end{figure}

%% file: prompts/IntentCoT.tex
\begin{figure}[ht]
\centering
\small
\begin{tcolorbox}[
  title=System Prompt for Intent CoT Query Generation,
  colframe=gray!70!black,
  colback=gray!5
]
You are an expert in user intent understanding. First, conduct user needs analysis, then provide \textbf{queries the user wants to ask}.

\vspace{1mm}
\textbf{\# Input Description} 
\begin{itemize}[itemsep=1pt, topsep=2pt]
    \item \textbf{[Session History]} $<$history$>$
    \item \textbf{[User Query]} $<$query$>$
    \item \textbf{[Agent Answer]} $<$response$>$
\end{itemize}

\textbf{\# Output Format}

\vspace{1mm}
{\footnotesize\ttfamily
\{\\
\hspace*{0.8em}"deeperIntent": \{\\
\hspace*{1.6em}"userSituation": "<user profile>",\\
\hspace*{1.6em}"motivations": "<latent intent>",\\
\hspace*{1.6em}"potentialIntents": ["<Intent 1>", ...]\\
\hspace*{0.8em}\},\\
\hspace*{0.8em}"response\_summary": \{\\
\hspace*{1.6em}"answeredInDetail": "<fully answered>",\\
\hspace*{1.6em}"answeredNotDetail": "<partially>",\\
\hspace*{1.6em}"notAnswered": "<not covered>"\\
\hspace*{0.8em}\},\\
\hspace*{0.8em}"recommend\_query": \{\\
\hspace*{1.6em}"query1": "<query>", // 20 chars\\
\hspace*{1.6em}"reason1": "<reason>", ...\\
\hspace*{0.8em}\}\\
\}
}

\vspace{1mm}
\textbf{\# recommend\_query\_rule}
\begin{enumerate}[itemsep=1pt, topsep=2pt]
    \item \textbf{Content Repetition Issue}: Recommended queries must not obviously repeat content from the user query or AI-Chat response;
    \item \dots (Additional rules omitted for brevity)
\end{enumerate}

\end{tcolorbox}
\caption{Abridged system prompt for intent-guided CoT query generation.}
\label{fig:intent_cot_prompt}
\end{figure}

%% file: prompts/LLMjudge.tex
\begin{figure}[ht]
\centering
\small
\begin{tcolorbox}[
  title=System Prompt for Rubric-based Evaluation,
  colframe=gray!70!black,
  colback=gray!5
]
\textbf{\# Role} \\
Senior Dialogue Recommendation Auditor. Evaluate whether recommended queries meet strict quality validation rules.

\vspace{1mm}
\textbf{\# Input Definitions} 
\begin{itemize}[itemsep=1pt, topsep=2pt, leftmargin=*]
    \item \textbf{Query Intent}, \textbf{Current Time}, \textbf{Session History}
    \item \textbf{User Query}, \textbf{Agent Answer}, \textbf{Recommended Query}
\end{itemize}

\textbf{\# Evaluation Rules} 
\begin{itemize}[itemsep=1pt, topsep=2pt, leftmargin=*]
    \item \textbf{R01 (Awkward Expression)}: Grammatical issues, improper collocations, punctuation errors.
    \item \textbf{R02 (Language Inconsistency)}: Must match the language of current user query.
    \item \textbf{R03 (Outdated Questions)}: No outdated content unless historical context requested.
    \item \textbf{R04 (Creative Content)}: For creative writing, use declarative/imperative only.
    \item \textbf{R05 (Irrelevant Needs)}: Avoid content unrelated to user query or agent answer.
    \item \dots (Additional rules)
\end{itemize}

\hrulefill

\textbf{\# Output Format} \\
JSON structure with two fields:
\begin{itemize}[itemsep=1pt, topsep=2pt, leftmargin=*]
    \item \texttt{is\_pass}: Boolean indicating pass/fail status
    \item \texttt{violations}: Array of violation objects containing \texttt{rule\_id}, \texttt{rule\_name}, and \texttt{reason}
\end{itemize}

\end{tcolorbox}
\caption{Abridged prompt for rubric-based quality assessment.}
\label{fig:rubric_eval_prompt}
\end{figure}